# X-ray Fluoroscopy Guided Localization and Steering of Medical Microrobots through Virtual Enhancement

Husnu Halid Alabay, Tuan-Anh Le, Hakan Ceylan*

Department of Physiology and Biomedical Engineering, Mayo Clinic College of Medicine and Science, 85259 Scottsdale, Arizona, United States

*Corresponding author's email address: Ceylan.Hakan@mayo.edu

## Abstract

In developing medical interventions using untethered milli- and microrobots, ensuring safety and effectiveness relies on robust methods for detection, real-time tracking, and precise localization within the body. However, the inherent non-transparency of the human body poses a significant obstacle, limiting robot detection primarily to specialized imaging systems such as X-ray fluoroscopy, which often lack crucial anatomical details. Consequently, the robot operator (human or machine) would encounter severe challenges in accurately determining the location of the robot and steering its motion. This study explores the feasibility of circumventing this challenge by creating a simulation environment that contains the precise digital replica (virtual twin) of a model microrobot operational workspace. Synchronizing coordinate systems between the virtual and real worlds and continuously integrating microrobot position data from the image stream into the virtual twin allows the microrobot operator to control navigation in the virtual world. We validate this concept by demonstrating the tracking and steering of a mobile magnetic robot in confined phantoms with high temporal resolution (< 100 ms, with an average of ~20 ms) visual feedback. Additionally, our object detection-based localization approach offers the potential to reduce overall patient exposure to X-ray doses during continuous microrobot tracking without compromising tracking accuracy. Ultimately, we address a critical gap in developing image-guided remote interventions with untethered medical microrobots, particularly for near-future applications in animal models and human patients.

## Introduction

Microrobots are emerging tiny mobile devices that have the potential to navigate and perform spatiotemporally precise tasks within confined, hard-to-access, and arbitrary spaces with remote human-in-the-loop control or semi-autonomously based on their locally responsive design(*1-3*). These capabilities can unlock numerous potential medical interventions inside the body that were once considered impractical or impossible in completely atraumatic ways (*4, 5*).

Over the past decade, various microrobot designs have been proposed for a wide variety of potential medical interventions, such as drug (*6*) and cell delivery (*7, 8*), tissue engineering, embolization (*9*), clot removal (*10*), etc. However, most of these applications have been demonstrated using in vitro settings, and the localization and navigation tasks of microrobots were accomplished under plain camera or optical microscopy. However, human/ or other animal bodies are mainly non-transparent to visible and near-infrared photons, ruling out the possibility of studying and developing optical imaging-guided medical interventions in most clinical scenarios considered thus far. As a result, a gradual effort has been shifting to employing medical imaging modalities for the detection and tracking of microrobots (*5, 9, 11-13*).

In developing a medical intervention with a microrobot system, real-time or near-real-time image acquisition is essential for localization, tracking the microrobot, and steering motion within the body. This critical safety and efficacy aspect is particularly vital for applications such as endovascular and neuro interventions, where safety standards are highest due to the low margin of error that can be tolerated. The available medical imaging systems that can stream data for continuous microrobot tracking with high spatial resolution and high-contrast images are currently limited to X-ray fluoroscopy or computed tomography (CT), positron emission tomography (PET), and ultrasonography (US). The US is limited to specific superficial body sites due to low penetration (3-4 cm) and suffers from strong scattering from bones and air pockets (*11*). X-ray fluoroscopy or CT and PET can localize microrobots deep in the body; however, these systems do not provide anatomical details about the robot's workspace.

In the context of X-ray fluoroscopy-guided endovascular catheter localization, for example, the occasional use of contrast enhancers has been a common practice (*14*). However, the continuous injection of these enhancers presents practical limitations, mainly due to the risk of nephrotoxicity (*15*). As a potential solution, intermittent infusion of contrast enhancers can be logically considered. Nevertheless, this approach would re-create the same challenge: the inability of external microrobot operators to accurately determine the microrobot's location and control its movements continuously.

Another approach to address a similar challenge in diagnostic imaging, PET/ magnetic resonance imaging (MRI), SPECT/ CT, and similar hybrid strategies were developed to overlay background anatomy with the target site (*16*). Nevertheless, developing a medical intervention inside such compact imaging platforms has insurmountable operational challenges around workspace sharing since these devices are built with closed bores. MRI can provide both imaging and navigation control for magnetic microrobots from those systems. Many procedures have been performed on patients in the MRI environment, including MRI-guided biopsy, radiotherapy, and thermal ablations (*17-19*). However, most untethered magnetic microrobots are not compatible with extremely high magnetic fields under MRI.

Consequently, real-time localization and navigation control represent a major barrier to the translational development of microrobots in both animal models and future human applications.

There is a compelling need to explore new concepts to move the field forward toward higher technological readiness level for innovative clinical applications (*5*).

Virtual reality environments are increasingly being used in medical education and diagnostic medicine for various purposes, including training and simulation, procedure planning, navigation, and guidance (*20*). Unlike most current medical imaging technologies, virtual reality (VR) can significantly enhance visual feedback and user intuitiveness, thereby improving medical interventions utilizing microrobot systems. VR can provide unobstructed 3D visualization and depth perception from a first-person perspective (*21*). It allows users to dynamically shift viewpoints, enabling complete 360° visualization, thus eliminating the need to abstract data from static 2D images. The ability to maneuver, scale, and rotate VR perspectives dynamically enhances internal spatial representation and improves intuitiveness, particularly for manipulating micro and nanoscale objects. Despite its outstanding advantages, VR-based techniques have not yet been explored for microrobot localization and navigation control.

This study addresses the challenges surrounding microrobot detection, localization, and steering under X-ray fluoroscopy guidance. Our proposed solution involves creating a simulation environment housing a precise digital replica, or virtual twin, of the microrobot's operational workspace. By synchronizing coordinate systems between the virtual and real worlds and continuously integrating microrobot position data from the image stream into the virtual twin, a human or machine operator can navigate it with near real-time temporal accuracy from the virtual world. We validate this concept by demonstrating the tracking and steering capabilities of a magnetic microrobot swimming in a confined, fluid-filled luminal phantom under X-ray fluoroscopy guidance. Our approach is versatile and applicable to live tissues, animal bodies, and human patients in a realistic clinical scenario. Beyond the confines of this study, virtual interfaces offer opportunities for incorporating various pre- and intraoperative data, including real-time physiology and haptic feedback. These integrations into the simulation environment have the potential to substantially enhance the safety and efficacy of microrobot interventions overall.

## Results and Discussion

***Identifying the strengths and limitations of remotely controlled microrobot navigation under X-ray fluoroscopy guidance***

We created an X-ray fluoroscopy-guided magnetic manipulation platform (eX-MMPt) for remote navigation control of microrobots, particularly in large animals and humans, as shown in Figure 1A i. It utilizes a clinical C-arm to capture fluoroscopic videos in cinefluoroscopic (cine) and digital subtraction (DS) modes with high temporal resolution without significant distortion or loss of information. This is crucial for guiding the magnetic manipulation of microrobots in real time. Also, the open workspace of C-arm allows the successful operation of a magnetic manipulation platform without compromising the imaging quality or requiring substantial re-design of two parallel working systems. In this study, our magnetic manipulation platform involves a seven-degrees-of-freedom robot arm. At the robot arm's end effector, we incorporated a motor that can rotate a permanent magnet to create rotating magnetic fields (Figure 1A ii). This configuration allows a computer-programmable control of the local magnetic field magnitude, magnetic field gradient, direction, and rotation frequency around a magnetic microrobot in a large volume that can accommodate a human subject (Figure 1A iii).

For the robotic navigation and localization demonstrations, we designed a magnetic, soft, helical-shaped miniature swimmer (HMS) (Figure 1Bi-ii). Analogous to our previous helical-shaped microrobots, HMS is magnetized in the direction proportional to its major helical axis for -torque-dominant cork-screw locomotion (*6, 7, 22-24*). Detailed information regarding the fabrication and magnetization of HMS is provided in Figure S1 and the Materials and Methods section. We placed HMS in a confined, fluid-filled two-dimensional luminal phantom to demonstrate and address detection, localization, and steering challenges arising from X-ray fluoroscopy.

An untethered microrobot intended for remote navigation within the human body under real-time X-ray fluoroscopy guidance must offer sufficient contrast to ensure reliable visibility. The quality of this contrast is crucial as it determines the total exposure dose received by the patient, thereby impacting potential short- and long-term side effects (*25*).

Figure 1B iii shows the visibility of HMS under C-arm. To semi-quantitatively assess the X-ray contrast of the microrobot, we utilized an iodine-based molecule, iohexol, and its intravascular Omnipaque® 350 formulation as the clinical standard. For real-time X-ray fluoroscopy-guided interventions, the medical device within the human body should provide contrast levels at least comparable to those of 30% Omnipaque® 350, which represents the acceptable minimum quality for real-time interventions within a human body (*26*). We found that the relative contrast of the HMS precursor composition (PDMS:NdFeB = 1:4 by weight) already surpasses that of 100% Omnipaque® 350, indicating that HMS would be readily visible within the human body (Figure 1C). This results from two main factors: (i) The presence of NdFeB magnetic particles, with an average size of approximately 5 μm, embedded within HMS, which are denser than iodine-containing molecules. (ii) Neodymium has a higher atomic number ($_{60}$Nd) than iodine ($_{53}$I), which typically results in better visibility with C-arm, due to higher X-ray attenuation.

As we applied magnetic actuation to HMS within the plastic phantom, we could conveniently capture the HMS and its locomotion in the cine and DS imaging modes of C-arm (Figure 1D, Supplementary Movie 1). In DS mode, however, there was a short delay until the robot emerged from its stationary position and became visible. This was due to the real-time digital subtraction of each frame from the first one by the C-arm computer. As soon as HMS starts making a displacement in the channel, it visually emerges from the blank background, leaving a white digital mark on its initial position. DS imaging mode is a valuable enabler for intravascular imaging of the chest and head due to the presence of bone (*27, 28*).

More importantly, we observed in both cine and DS modes that the lumen and the phantom are almost completely invisible to the external operator. The background also severely fluctuated by the presence of other high-contrast materials in the X-ray path, such as the robot arm or a big permanent magnet coming across the imaging space, reducing the overall reliability of the visibility for the structural details in the robot workspace. Without precise localization information in its workspace, the remote control of a microrobot becomes paralyzed, and steering becomes unreliable. In a medical intervention, this would constitute an unsurmountable safety barrier.

***Creation of the virtual interface***

To address the challenge of navigating HMS in limited environmental detail under C-arm guidance, we explored the possibility of creating a virtual replica of the phantom testbed and continuously synchronizing the position data of HMS within the simulated environment as it moves within the physical phantom. In our system, we establish a connection between the real environment setup and the virtual environment setup using our custom interface. Each system performs its unique tasks and shares results with each other via the Robot Operating System (ROS) network, ensuring a flexible design and structure. In future studies, we envision the possibility of integrating different systems and devices into our setup to enhance its functionality. Figure 2A illustrates our data handling process among the various systems within this study and outlines potential future setups to improve our system for future applications.

Our approach involves initially developing a training-based object detection algorithm for HMS (See in detail in Microrobot detection model of Materials and Methods section). This algorithm enables us to retrieve various unique identifier data about the HMS, including the confidence value of the detection success rate, a bounding box surrounding the HMS body, the geometric center of the HMS, and the position data of the geometric center within the image frame (Figure 2Bi). As a result, we can match these identifiers between real and virtual world (Figure 2Bii). This data is subsequently sent into our simulation environment (Unity®) using ROS2 (*29*) as our network communication system, which works under a customized Phyton 3 script. For simulation purposes, there are several options which present 3D simulations environment and data transfer. Systems such as Gazebo, an open-source 3D robotics simulator, or CoppeliaSim, formerly known as V-REP robot simulator, can be used for this kind of purposes. They can offer some unique features such as Gazebo's direct connectivity with ROS interface, but Unity® environment preferred in this study due to its allowance for augmented and virtual reality applications (*30*). Further, it is also possible to connect Unity® with other Unity® based applications, simulations, and devices in the future which can enhance our system's performance and capabilities.

Second, we developed a spatial calibration algorithm that links the real-world dimension and video frame size to those in the virtual world before the navigation is started. Before each procedure, calibration between these two worlds should be done in order to have the same positional information between HMS and its environment in both worlds. To perform this calibration, either we can use the location of two end points of our phantom on our initial image, or we can place fiducials to mark these points if they are not clearly visible due to their x-ray contrasts (Figure S2A). We should combine these positional data with the real length of the phantom to define their unit conversion. To perform the calibration, first, we use initial frame from our input video which shows our whole phantom model, and fiducials if they are available, within our experiment setup. Then, we mark two end points (Figure S2A) of the phantom, or end points of the fiducials, on the image and save their pixel locations. To be able to match their position, we calculate our pixel-to-mm conversion by using pixel distance between these two points and real length of the phantom. After that, we place the virtual phantom on the first chosen point and rotate it until it reaches to the second chosen point's location. So, by doing this operation, we can match position and rotation of the real phantom and its virtual twin.

To validate precise one-to-one matching between HMS in the real and virtual worlds, we first recorded nearly two minutes of navigation video by using a Red-Green-Blue (RGB) camera. We processed each frame within the video, generated the identifier position data, and transmitted it to the simulation environment via our communication interface one-by-one. By processing each frame like that, we can create a data flow of each frame's data to mimic the real-time twin of the experiment in a simulation environment. With the calculations in the background, the slight time difference between the real-world and virtual-world data has always been below 100 ms, typically as low as ~20 ms, is negligible and below human detection capacity, owing to the high frame rate of our input video.

Unity®'s visual interface allows the display of HMS derived data, such as its velocity. As we move HMS, we can compute the instantaneous velocity corresponding to the displacement between each consecutive frame analyzed. Additionally, we calculate the 1-second integrated velocity and the 10-second integrated velocity, thereby enhancing the reliability of the velocity data, smoothen out fluctuations and noise in the velocity data.

We can accurately predict the actual position and successfully steer HMS by integrating real-time microrobot displacement data from an RGB camera or C-arm into the virtual twin (Figure 2C-Figure S2B, Supplementary Movie 2).

### *Virtually enhanced real-time localization and navigation control*

To demonstrate the feasibility of real-time navigation of HMS under X-ray fluoroscopy-mediated detection in the virtual interface, we placed a camera in front of the C-arm tower screen and streamed the image data directly to our main computer (Figure S3). We execute our object detection, calibration, and localization algorithms by processing the live image stream and sending the identifier data to the simulation environment. Figure 3A and Supplementary Movie 3 demonstrate the effectiveness of the HMS navigation in the simulation interface with *ca.* 20-25 ms delay in the virtual interface. Strikingly, we could conveniently steer HMS into the side branch without missing the branching.

To further demonstrate the effectiveness of localization and steering in the virtual interface, we played a maze game with a 5 mm NdFeB sphere bead. The maze was completely closed and consisted of one *START*, one *END* point, and multiple dead ends (Figure 3B i-ii). We created the virtual twin of the maze, calibrating its coordinate systems and dimensions with the physical maze using X-ray visible fiducials. Then, we successfully moved the bead through the maze in seven steps in the virtual interface to the exit (Figure 3B iii -iv, Supplementary Movie 3).

***Envisioning virtual anatomy enhancement in a future microrobotic intervention***

Our method has significant implications for advancing the translational development of microrobot systems. To improve visual feedback and user intuitiveness, we envision constructing the HMS workspace and anatomical details through high-resolution MRI or CT scans (Figure 4). Subsequently, this comprehensive anatomical data can be seamlessly integrated into the virtual reality environment. This virtual twin is created in the preoperative preparation stage of the micro robotic intervention task, following matching the coordinates with the real patient positioning and orientation through assistive fiducials. The creation of this virtual twin occurs during the preoperative preparation stage of the micro robotic intervention task, where coordinates are matched with real patient positioning and orientation using assistive fiducials. This method facilitates real-time detection and tracking of the microrobot under X-ray fluoroscopy or similar streaming medical imaging modality and guidance within an immersive, user-friendly virtual interface. We consider that this integrated approach will be an essential enabler for ensuring the safety and efficacy standards demanded by these emerging technologies in realistic clinical applications within the body.

While C-arm-guided procedures are widely utilized in clinical settings, a significant concern associated with X-ray fluoroscopy is the exposure to ionizing radiation, which can have both deterministic and stochastic effects. Our approach, utilizing a virtual anatomical enhancement strategy, aims to mitigate this risk by enabling the reduction of pulse rates for cinefluography acquisition. This strategy focuses exclusively on detecting the high contrast microrobot, thereby substantially decreasing the exposure of both patients and physicians to X-ray radiation doses.

As a moving concept, digital twins are becoming more than a static virtual copy of human organs and tissues; they become evolving models that can reflect present changes and predict future events. In this regard, we envision that integrating medical microrobot and digital twin interfaces is a strategic alliance to move the field forward in translational development. Further opportunities in virtual augmentation to a medical microrobot operation may involve patient-specific, dynamic models, which can dramatically enhance dexterous microrobot control. Virtual interfaces can pave the way for fully autonomous interventions, incorporating various intraoperative sensory data from the patients, periodic or atypical body motions, and haptic feedback into the virtual twin environment to ensure the safety and efficacy of the intervention. Those motions cannot be captured using hybrid imaging modes and properly managed during the operation. incorporating complex body motions in real-time into the virtual realm that integrates multiple sensory data can profoundly enhance the potential safety concerns associated with microrobot deployment within sensitive body sites.

**Conclusion**

This proof-of-concept study lays the groundwork for real-time detection, tracking, and navigational control of untethered microrobots inside large animals and human body by utilizing a virtual twin interface. It demonstrates the navigation of model magnetic robots within a static, 2D phantom testbed. While X-ray fluoroscopy is powerful for microrobot detection and offers exceptional capability for real-time imaging deep within the body, its primary limitation lies in producing 2D images. Future investigations in this direction will concentrate on microrobot control in a 3D workspace within virtual twin interfaces while using 2D fluoroscopy images.

**Materials and Method**

*Materials:* Unless described otherwise in its relevant context, all chemicals were purchased and used from Sigma Aldrich in high purity.

*Magnetic field control:* An LBR Med 7 R800 robot arm (KUKA) with seven degrees of freedom was employed to control the position and orientation control of the global magnetic field in the workspace of the microrobot. Our KUKA robot arm's software version is V1.5.4-2 and KUKA Robot arm programmed with its unique application (KUKA Sunrise Workbench – 2.6.5_6). Attached to the robot arm end effector, a custom-designed DC motor (Maxon)-run part facilitates the high-performance rotation of an external permanent magnet (K&J Magnetics) up to ~100 Hz.

*X-ray fluoroscopy imaging:* X-ray images were captured in cinefluoroscopic and digital subtraction angiographic modes using an OEC 3D C-arm (General Electric, Boston, MA). The DICOM image files from the C-arm were converted to .JPEG format and .MP4 using RadiAnt DICOM Viewer (version 2023.1, 64-bit) software.

*Camera imaging:* Video recordings in Figure 2 were captured using a Blackfly S Sony IMX546 2/3" CMOS camera. For real-time magnetic navigation system with the C-arm described in Figure 3, a Logitech webcam camera (Logitech Brio 4K webcam V-U0040), with 640x480 frame size, was localized in front of the C-arm tower monitor to capture and transfer the image data to Unity® (Figure S3B).

*Fabrication of HMS:* The fabrication of HMS involved designing a master positive mold and a pin with predefined dimensions using Fusion 360 (Autodesk, San Francisco, California), followed by their 3D printing using a commercial stereolithography (SLA) printer (FormLabs 3B+, FormLabs, Somerville, MA). Subsequently, the master positive mold and pin were utilized to create a negative mold made from silicon elastomer. A mixture with a 10:1 mass ratio of SYLGARD™ 184 Silicone Elastomer Base and Curing Agent (DOW Chemical Company) was prepared and cured in a heater at 80°C for 2 hours. Prior to the subsequent step, the surface of the negative mold was passivated using plasma treatment and alcohol passivation, as described previously (*31*). The silicon elastomer: NdFeB powder from Magnequench (D50 = 5 $\mu$m) mass ratio was 1:4. The thoroughly mixed precursor composite was then degassed under vacuum and subsequently cast into the passivated negative PDMS mold and cured in the heater at 80°C for 2 hours. Following the curing, the helical swimmer was gently demolded. Finally, the HMS was magnetized in a direction perpendicular to its major helical axis using our custom-made magnetic yoke under 1.1 T. The magnetization protocol is described in detail in Figure S1.

*Real-world models:* Phantom models and their assistive parts with pre-defined dimensions were designed using Fusion 360. The transparent phantom with fluid-filled closed channel in Figure 1-3 was designed with a length of 186 mm and a width of 12 mm. To illustrate the navigation capabilities of HMS through branches of varying sizes, a phantom with four channels was meticulously constructed. The initial channel maintains a consistent diameter of 6 mm, while the remaining three channels adopt a conical structure, commencing with diameters of 6 mm and gradually narrowing to 1 mm, 2 mm, and 3 mm, respectively. A sacrificial mold was 3D printed using a K1 Mx AI fast 3D printer (Creality, Shenzhen, China) using acrylonitrile butadiene styrene (ABS) filament (Creality). This mold is then sprayed with non-adhesive ease release 200 spray (Mann release technologies, Macungie, PA, USA) and placed in a container. Then, the container was filled with SYLGARD™ 184 Silicone Elastomer Kit, where components A and B were mixed in a 1:10 ratio. Following overnight curing at room temperature, the sacrificial ABS mold was dissolved with acetone the next day, resulting in the phantom in its final form (Figure S4). The maze was 3D printed from the ABS filament, fully closed and opaque, to demonstrate real-time navigation control in the virtual interface.

*Virtual twins:* The virtual twins of the phantom testbeds were created by importing the .OBJ file of the phantom design (Fusion 360) into the Unity® 2022.3.0f1 editor version (Unity® Technologies).

*Spatial calibration:* To achieve spatial alignment of the phantom testbeds and virtual twins, a calibration method was developed. Prior to the navigation task, an initial image was captured to (i) match the frame size of the input video to the virtual frame size and correct (ii) position and (iii) rotation of the object within the frame to the virtual twin. X-ray-opaque fiducials were used to identify the starting and ending positions of the phantoms defined on the image, followed by pixel-to-mm conversion. To execute this task, a function was added to the Python and C# codes. This function uses the initial image to register the fiducials. The first fiducial was used to match starting points of real phantom and virtual twin and second was used to match their rotation around the starting point. Subsequently, Unity® positions the virtual twin of the phantom by adjusting its position and angle to match accurately the real-world positioning.

*Communication network:* A new version of the Robot Operating system, ROS2, more specifically ROS2 Humble, was used to create a communication network for all separate system works within our project. Everything works within our local network and each system can send and retrieve data by publishing and subscribing to the related topics, data packages, on ROS2 network. In our work, we have python codes for object detection and ROS2 communication and C# codes for Unity® connection and simulation environment control. For future uses, everything can be connected to the internet network for untethered communication between each system which can lead cross-continental connection. After the spatial calibration, the connection of the Unity® system with the ROS2 network was established using ROS-TCP-Endpoint libraries from Unity® Technologies. With this connection, detected HMS data could be received, and its virtual twin position could be adjusted in real time. As shown in Figure 2B, their positions match accurately without any detectable mismatching in the given pixel size.

*Microrobot detection model:* A dataset consisting of 318 images of HMS was created from the videos of microrobot navigation within the phantoms, captured by an RGB camera or C-arm, to train the object detection model. Roboflow platform (Roboflow, Inc, Des Moines, Iowa, USA) was used to create and annotate the database (*32*). The built-in data augmentation functions in Roboflow were also utilized to expand the microrobot image dataset for training, creating new iterated images of the originals with various defects and changes such as flip, rotation, blur, saturation, brightness, exposure, and grayscale. To facilitate testing of our model's success, validation and test sets were needed; thus, this expansion was applied after separating some images for these sets to preserve the original form of our detection target. Consequently, a total of 1210 images were obtained, with 92% (1115 images) in the *Train set*, 5% (63 images) in the *Validation set*, and 3% (32 images) in the *Test set*. Following dataset preparation, a new version of the You Only Look Once (YOLOv8, Ultralytics, MD, USA) algorithm was employed for object detection (*33*). The model underwent training for 300 iterations to reduce object losses and missed detections. Upon completion of training, a weight file containing all the learned detections from the training was generated and made ready for use in our system. Subsequently, a Python script was developed to utilize the trained dataset for microrobot detection on pre-recorded videos or real-time footage. In the recreation of the twin HMS in Unity®, a detected robot within the imaging frame was outlined with a bounding box, and its center point was marked with a circle in the background. This HMS position data was sent to Unity® over the ROS2 network. For the visual validation of a detected HMS, a bounding box was also printed in the recorded videos around HMS, as shown in Figure 2 and Figure S2, whenever a detection event occurred.

**Supplementary Materials**

**Acknowledgments**

Funding: This research was mainly funded by the startup funding provided by Mayo Clinic for H.C and partly funded by American Heart Association Career Development Award (grant number 23CDA1040585). The authors acknowledge Hassan Albadawi, MD, for support with the C-arm operation.

Competing interests: The authors declare that they have no competing interests.

Data and materials availability: All data needed to evaluate the conclusions in the paper are presented in the paper and/or the Supplementary Materials.

**References**

1. B. J. Nelson, I. K. Kaliakatsos, J. J. Abbott, Microrobots for Minimally Invasive Medicine. *Annual Review of Biomedical Engineering* **12**, 55-85 (2010).
2. M. Sitti *et al.*, Biomedical Applications of Untethered Mobile Milli/Microrobots. *Proceedings of the IEEE* **103**, 205-224 (2015).
3. P. E. Dupont *et al.*, A decade retrospective of medical robotics research from 2010 to 2020. *Science Robotics* **6**, eabi8017 (2021).

4.  H. Ceylan, J. Giltinan, K. Kozielski, M. Sitti, Mobile microrobots for bioengineering applications. *Lab on a Chip* **17**, 1705-1724 (2017).
5.  H. Ceylan, I. C. Yasa, U. Kilic, W. Hu, M. Sitti, Translational prospects of untethered medical microrobots. *Progress in Biomedical Engineering* **1**, 012002 (2019).
6.  H. Ceylan *et al.*, 3D-Printed Biodegradable Microswimmer for Theranostic Cargo Delivery and Release. *ACS Nano* **13**, 3353-3362 (2019).
7.  I. C. Yasa, A. F. Tabak, O. Yasa, H. Ceylan, M. Sitti, 3D-Printed Microrobotic Transporters with Recapitulated Stem Cell Niche for Programmable and Active Cell Delivery. *Advanced Functional Materials* **29**, 1808992 (2019).
8.  N. O. Dogan *et al.*, Remotely Guided Immunobots Engaged in Anti-Tumorigenic Phenotypes for Targeted Cancer Immunotherapy. *Small* **18**, 2204016 (2022).
9.  X. Liu *et al.*, Magnetic soft microfiberbots for robotic embolization. *Sci Robot* **9**, eadh2479 (2024).
10. I. S. M. Khalil *et al.*, Mechanical Rubbing of Blood Clots Using Helical Robots Under Ultrasound Guidance. *Ieee Robotics and Automation Letters* **3**, 1112-1119 (2018).
11. S. Pané *et al.*, Imaging Technologies for Biomedical Micro- and Nanoswimmers. *Advanced Materials Technologies* **4**, 1800575 (2018).
12. A. Aziz *et al.*, Medical Imaging of Microrobots: Toward In Vivo Applications. *ACS Nano* **14**, 10865-10893 (2020).
13. V. Iacovacci *et al.*, High-Resolution SPECT Imaging of Stimuli-Responsive Soft Microrobots. *Small* **15**, e1900709 (2019).
14. A. Ramadani *et al.*, A survey of catheter tracking concepts and methodologies. *Med Image Anal* **82**, 102584 (2022).
15. Q. A. Rao, J. H. Newhouse, Risk of nephropathy after intravenous administration of contrast material: a critical literature analysis. *Radiology* **239**, 392-397 (2006).
16. S. D. Voss, SPECT/CT, PET/CT and PET/MRI: oncologic and infectious applications and protocol considerations. *Pediatr Radiol* **53**, 1443-1453 (2023).
17. T. Penzkofer *et al.*, Transperineal in-bore 3-T MR imaging-guided prostate biopsy: a prospective clinical observational study. *Radiology* **274**, 170-180 (2015).
18. M. A. van den Bosch *et al.*, MRI-guided needle localization of suspicious breast lesions: results of a freehand technique. *Eur Radiol* **16**, 1811-1817 (2006).
19. M. van den Bosch *et al.*, MRI-guided radiofrequency ablation of breast cancer: preliminary clinical experience. *J Magn Reson Imaging* **27**, 204-208 (2008).
20. T. Sun, X. He, Z. Li, Digital twin in healthcare: Recent updates and challenges. *Digit Health* **9**, 20552076221149651 (2023).
21. Z. Zhang *et al.*, Active mechanical haptics with high-fidelity perceptions for immersive virtual reality. *Nature Machine Intelligence* **5**, 643-655 (2023).
22. Y.-W. Lee, H. Ceylan, I. C. Yasa, U. Kilic, M. Sitti, 3D-Printed Multi-Stimuli-Responsive Mobile Micromachines. *ACS Applied Materials & Interfaces* **13**, 12759-12766 (2021).
23. I. C. Yasa, H. Ceylan, U. Bozuyuk, A.-M. Wild, M. Sitti, Elucidating the interaction dynamics between microswimmer body and immune system for medical microrobots. *Science Robotics* **5**, eaaz3867 (2020).
24. X. Hu *et al.*, Magnetic soft micromachines made of linked microactuator networks. *Science Advances* **7**, eabe8436 (2021).

25. D. Vanzant, J. Mukhdomi, in *StatPearls*. (StatPearls Publishing Copyright © 2024, StatPearls Publishing LLC., Treasure Island (FL), 2024).
26. Z. Zhang *et al.*, Treatment of Ruptured and Nonruptured Aneurysms Using a Semisolid Iodinated Embolic Agent. *Advanced Materials* **34**, 2108266 (2022).
27. T. F. Meaney *et al.*, Digital subtraction angiography of the human cardiovascular system. *American Journal of Roentgenology* **135**, 1153-1160 (1980).
28. J. R. Little, A. J. Furlan, M. T. Modic, B. Bryerton, M. A. Weinstein, Intravenous digital subtraction angiography in brain ischemia. *JAMA* **247**, 3213-3216 (1982).
29. S. Macenski, T. Foote, B. Gerkey, C. Lalancette, W. Woodall, Robot Operating System 2: Design, architecture, and uses in the wild. *Science Robotics* **7**, eabm6074 (2022).
30. M. S. P. d. Melo, J. G. d. S. Neto, P. J. L. d. Silva, J. M. X. N. Teixeira, V. Teichrieb, in *2019 21st Symposium on Virtual and Augmented Reality (SVR)*. (2019), pp. 242-251.
31. S.-H. Kim, S. Lee, D. Ahn, J. Y. Park, PDMS double casting method enabled by plasma treatment and alcohol passivation. *Sensors and Actuators B: Chemical* **293**, 115-121 (2019).
32. B. Dwyer, Nelson, J., Hansen, T., et. al. , Roboflow (Version 1.0) [Software]. *Available from https://roboflow.com. computer vision.*,  (2024).
33. G. Jocher, Chaurasia, A., & Qiu, J. (2023). Ultralytics YOLO (Version 8.0.0).[Software]. Available from https://github.com/ultralytics/ultralytics. (2023).

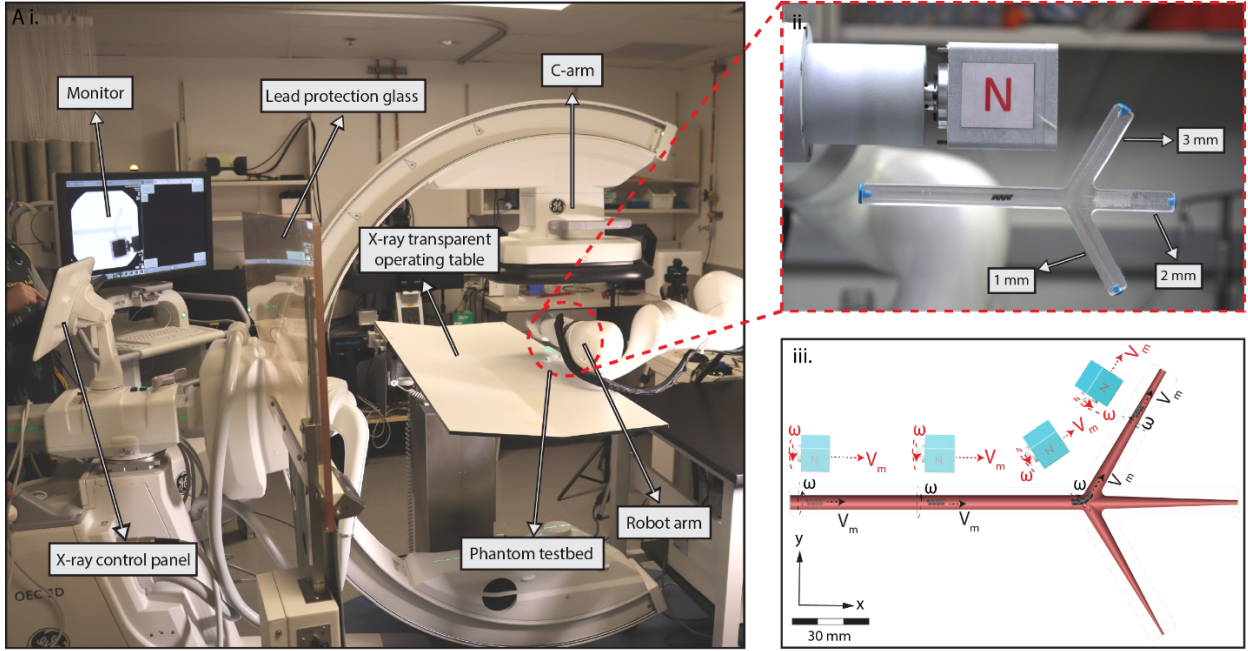

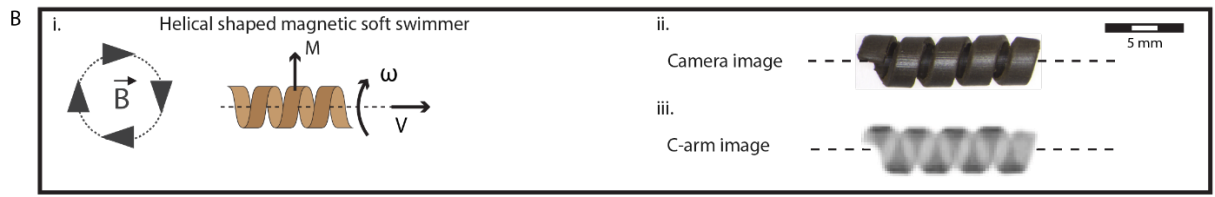

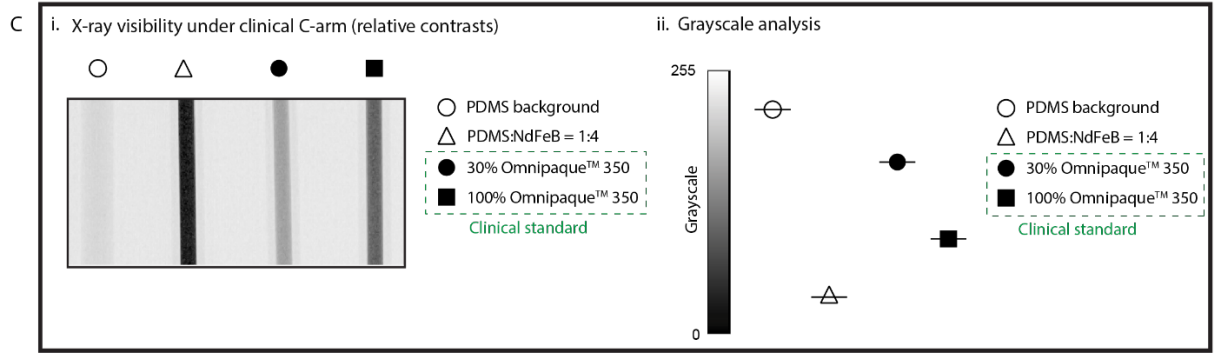

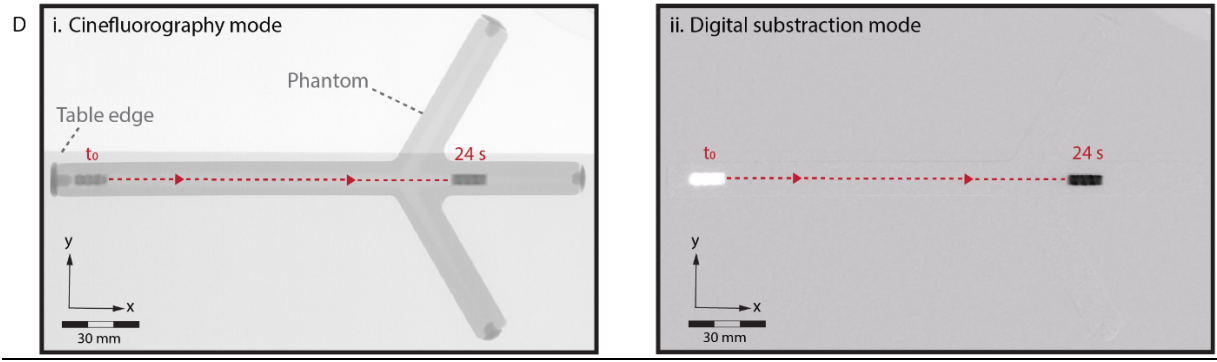

**Figure 1. Limitations in the real-time localization and navigation of miniature robots under real-time X-ray fluoroscopy guidance.** (A) (i) Our eX-MMPt system. For the remote magnetic manipulation, (ii) a permanent magnet is connected to a DC motor to create rotating magnetic fields, where phantom model has multiple ends with different dimension. (iii) The direction and magnitude of the magnetic fields in the local 3D workspace around the miniature robot are programmable with a seven-degree-of-freedom robot arm by controlling the position and orientation of the permanent magnet. (B) (i) A model helical-shaped miniature swimmer (HMS) that moves under rotating magnetic fields with cork-screw locomotion. (ii), (iii) The camera and X-ray fluoroscopy images of HMS. (C) (i) Relative X-ray contrast of HMS precursor composite inside 1mL syringe with respect to Omnipaque™ 350 filled syringes. 30% Omnipaque™ 350 sets the lower boundary of iodine concentration that offer visibility within the human body. (ii) using C-arm greyscale analysis of the relative X-ray contrasts showing HMS is reliably detectable for real-time C-arm guided medical interventions inside the human body, as its relative x-ray contrast surpass the clinical standard. (D) Visualization of HMS in the (i) cinefluorography and (ii) digital subtraction modes as it moves inside a fluid-filled phantom channel, demonstrating the poor anatomical detail of the environment in which the robot is navigated.

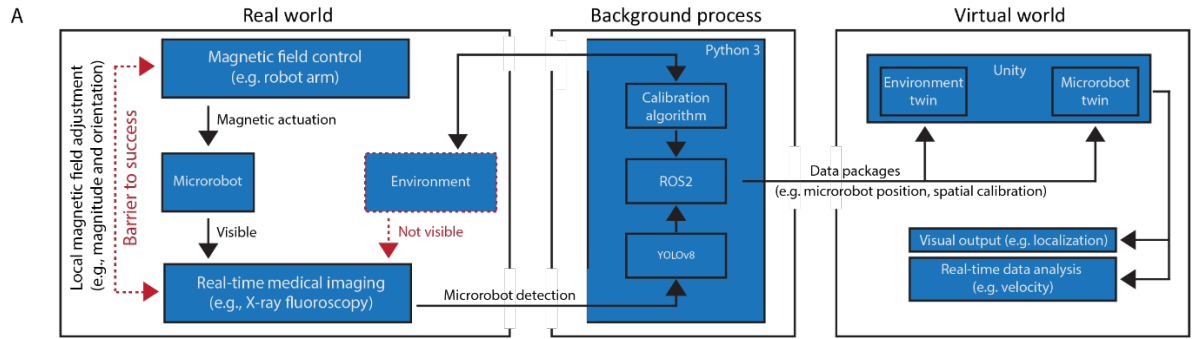

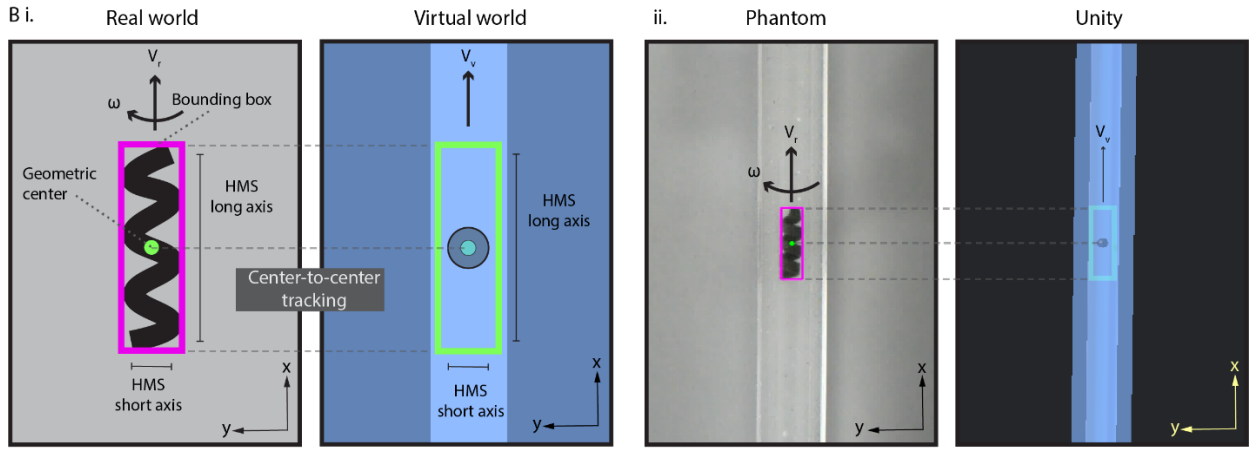

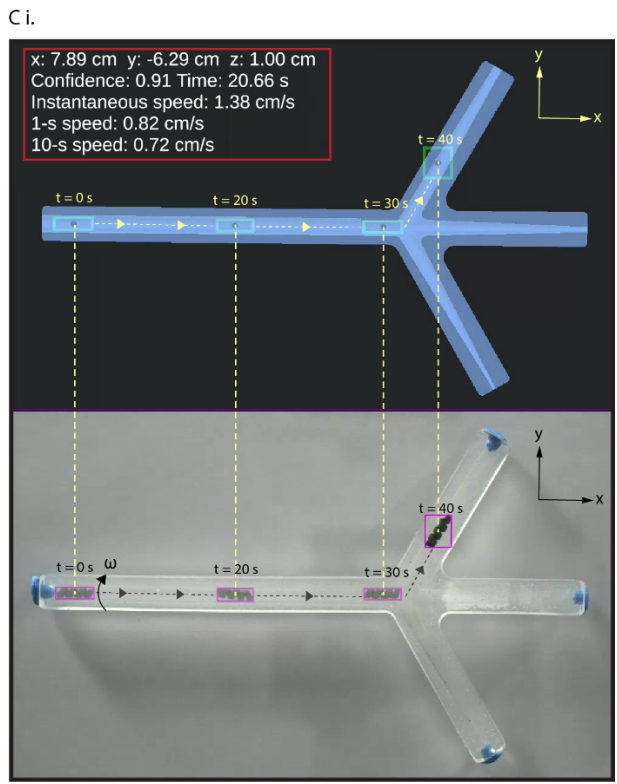 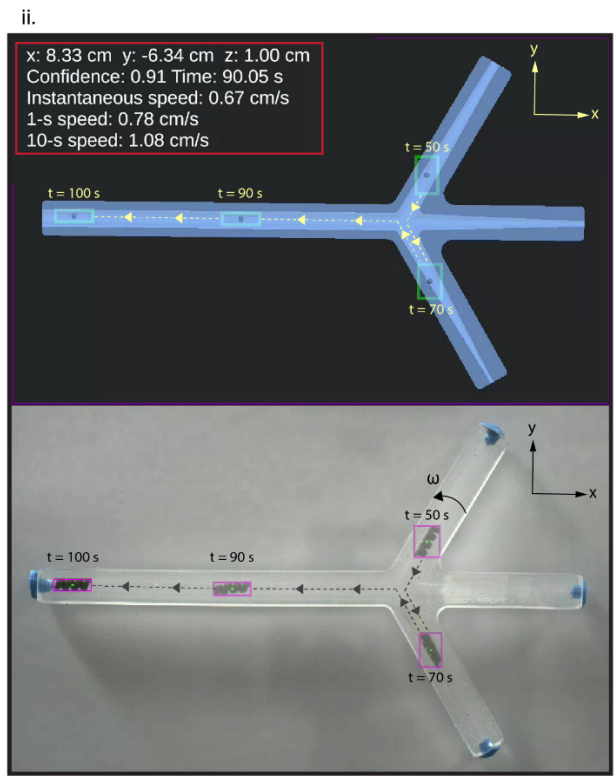

**Figure 2. Creation and validation of the virtual interface.** (A) The schematic diagram illustrating the data process and interface connections in the proposed virtual enhancement strategy. (B) (i) An illustration of our approach for HMS detection by the detection algorithm, with a bounding box created around it. A replicate of the bounding box is created as the virtual twin of the HMS in Unity®, and the position of its geometric center is used to localize it in the virtual twin of the phantom. (ii) Demonstrating synchronized detection and localization of HMS in real and virtual worlds. (C) Validating the accurate spatial and temporal synchronization of an HMS navigation in an RGB camera video in the virtual environments with time stamps, position, and HMS speed data (i) First half of the movement from left to right (0s-40s), (ii) second half of the movement from right to left (50s-100s).

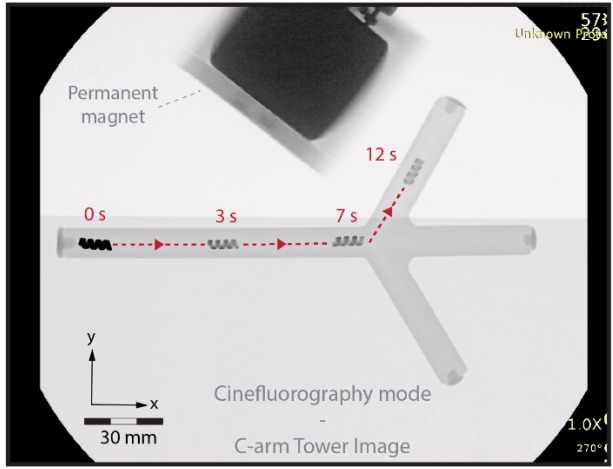
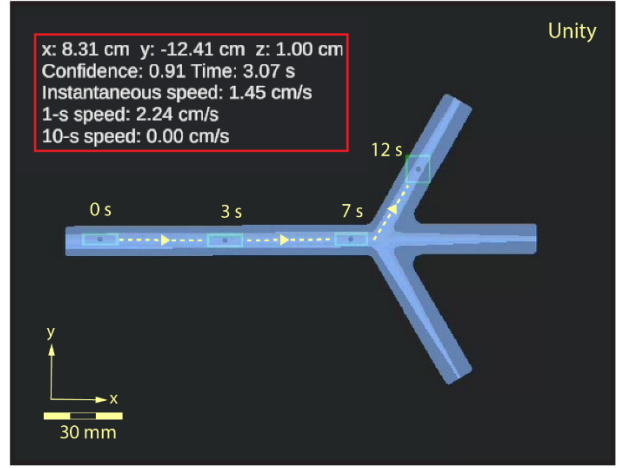
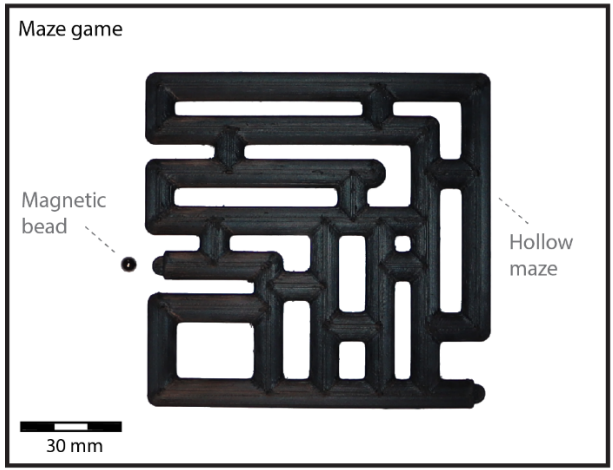
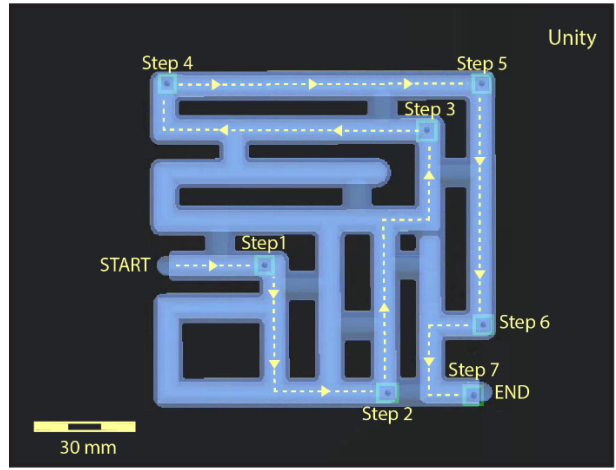
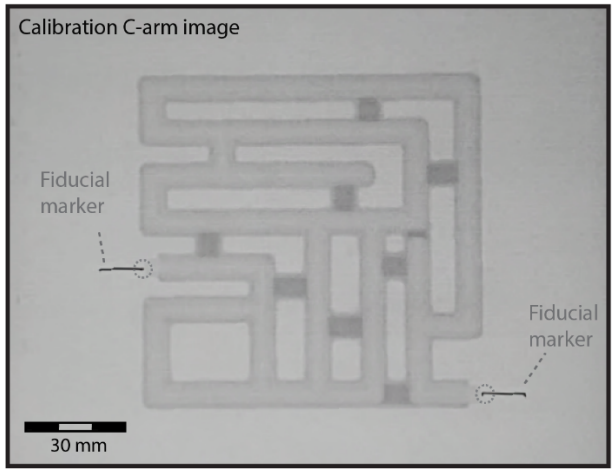
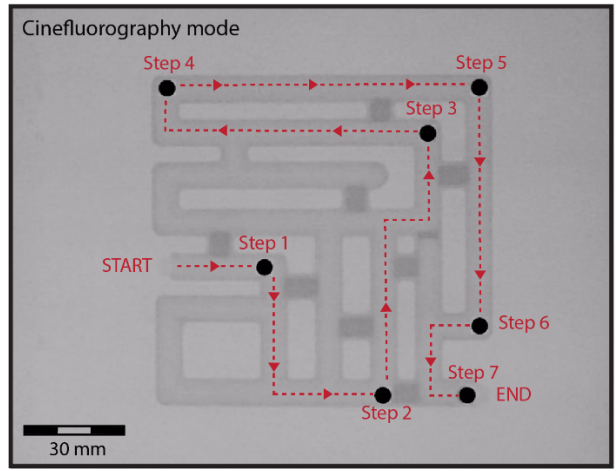

**Figure 3. X-ray fluoroscopy-guided real-time localization and telerobotic navigation control in the virtual twin interface.** (A) (i) Real-time validation of HMS navigation. As the magnet and part of the robot arm move within the imaging field, the relative contrast of HMS and the background phantom fluctuates, leading to unreliable robot control. (ii) Spatiotemporally precise steering control of HMS in the virtual twin, together with real-time derived data (e.g., speed). (B) Solving a maze game by tracking and steering a magnetic bead in the virtual interface. (i) RGB camera image of a closed maze and a NdFeB magnet bead. (ii) The first X-ray snapshot of the maze featuring fiducial markers calibrates the coordinates between the virtual and real worlds before the telerobotic navigation starts. (iii) Tracking and navigating the bead inside the virtual maze twin from *START* to *END* with stepwise robot arm control of the remote magnetic actuation. (iv) Validation of the magnetic bead localization in the maze under the C-arm.

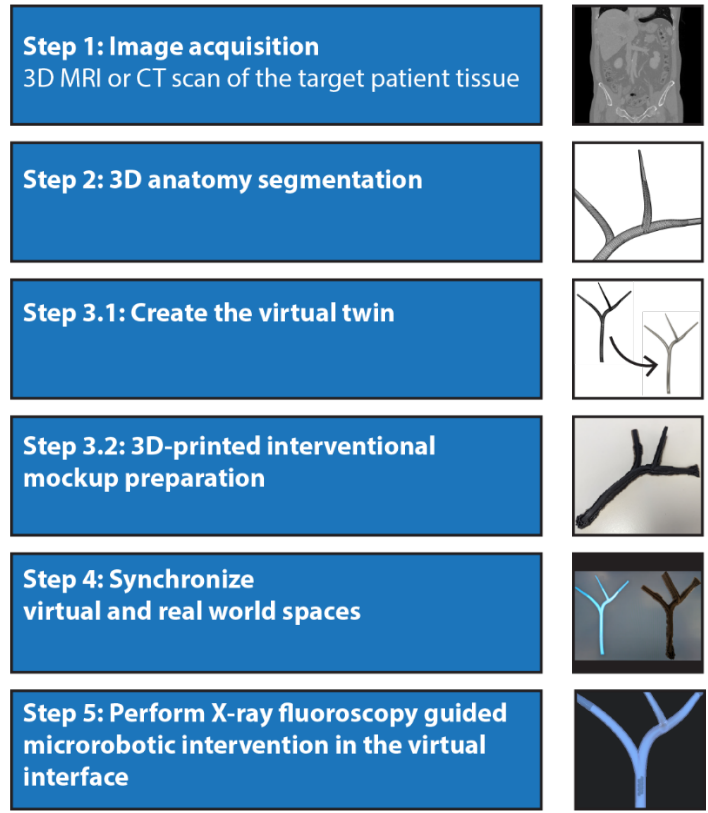

**Figure 4. Envisioned clinical integration process.** Virtual anatomy enhancement is essential for real-time localization and precise remote navigation control of miniature robots within the human body with high precision and acceptable safety. To implement this strategy in a future medical intervention, a preoperative planning stage would be needed. This stage would begin with a high-resolution scanning of the target tissue or organ anatomy using 3D imaging systems like MRI or CT. Following this, the segmentation of the image data would be used to create a digital twin of the anatomical details in the virtual environment. The segmentation data could be further utilized to create 3D-printed models and provide an interventional or surgical mock-up for preoperative study. Once the coordinate systems of the virtual and real environments were synchronized, the microrobotic intervention would be safely performed with the visual tracking and localization feedback integrated with the virtual interface.